\documentclass[11pt,a4paper]{article}
\usepackage[hyperref]{emnlp-ijcnlp-2019}
\usepackage{times}
\usepackage{latexsym}

\usepackage{amsmath}
\usepackage{graphicx}
\usepackage{multirow}
\usepackage{booktabs}
\usepackage{algorithm}
\usepackage{algorithmic}

\usepackage{url}
\usepackage{todonotes}

\usepackage[font=small]{caption}

\aclfinalcopy 

\title{Meta Reasoning over Knowledge Graphs}

\author{
 Hong Wang$^\dagger$,
 Wenhan Xiong$^\dagger$,
 Mo Yu$^{\ddag \ast}$, 
 Xiaoxiao Guo$^{\ddag \ast}$, 
 Shiyu Chang$^\ddag$, 
 William Yang Wang$^\dagger$
\\
 $^\dagger$ University of California, Santa Barbara\\
 $^\ddag$ IBM Research\\
 \{hongwang600, xwhan, william\}@cs.ucsb.edu, yum@us.ibm.com, \{xiaoxiao.guo, shiyu.chang\}@ibm.com  
 }

\date{}

\begin{document}
\maketitle

\begin{abstract}

 The ability to reason over learned knowledge is an innate ability for humans and humans can easily master new reasoning rules with only a few demonstrations. While most existing studies on knowledge graph (KG) reasoning assume enough training examples, we study the challenging and practical problem of few-shot knowledge graph reasoning under the paradigm of meta-learning. We propose a new meta learning framework that effectively utilizes the task-specific meta information such as local graph neighbors and reasoning paths in KGs. Specifically, we design a meta-encoder that encodes the meta information into task-specific initialization parameters for different tasks. This allows our reasoning module to have diverse starting points when learning to reason over different relations, which is expected to better fit the target task. On two few-shot knowledge base completion benchmarks, we show that the augmented task-specific meta-encoder yields much better initial point than MAML and outperforms several few-shot learning baselines.
 

\end{abstract}

\section{Introduction}
Knowledge Graphs \cite{DBpedia,freebase,wikidata} represent entities' relational knowledge in the form of triples, i.e., (subject, predicate, object), and has been proven to be essential and helpful in various downstream applications such as question answering \cite{freebaseQA, LargeScaleSinpleQuestion,KBQA_yih,kbqa_mo17}. Since most existing KGs are highly incomplete, a lot of studies~\cite{imt_BordesUGWY13,imt_TrouillonWRGB16,randomWalk_Lao} have been done in automatically completing KGs, i.e., inferring missing triples. However, most of these studies only focus on frequent relations and ignore the relations with limited training samples. As a matter of fact, a large portion of KG relations are actually long-tail, i.e., they have very few instances. Therefore, it is important to consider the task of knowledge graph completion under few-shot learning setting, where limited instances are available for new tasks. \citet{one-shot-KBR} first propose a graph network based metric-learning framework for this problem but the metric is learned upon graph embeddings and their method does not provide reasoning rationales for the predictions. 

In contrast, we propose a meta reasoning agent that learns to make predictions along with multi-hop reasoning chains, thus the prediction of our model is fully explainable. In this problem setting, each task corresponds to a particular relation and the goal is to infer the end entity given the start entity (i.e., the query). 
Following the recent work~\cite{few-shot-miniimagenet,maml,MAML_translation,MAML_SQG,relationNet,meta-learning-mishra} on meta-learning, we aim to learns a reasoning agent that can effectively adapt to new relations with only a few examples. This is quite challenging since the model must learn to leverage its prior learning experience for fast adaptation and at the same time avoid overfitting on the few-shot training examples. 
Model-agnostic meta-learning algorithm (MAML) \cite{maml} is a popular and general algorithm to solve this problem. It aims to learn an initial model that captures the common knowledge shared within the tasks so that it can adapt on the new task quickly.
But one problem of MAML is that it only learns a single initial model, which can not fit the new task without training, and has limited power in the case of diverse tasks \cite{few-shot-closer}. Another problem is that MAML only learns the common knowledge shared within the tasks without taking advantage of the relationship between them since no task-specific information is used when learning the initial model. 


In order to learn the relationship between tasks, the model must be aware of the identity of the current task, such as the query relation in our problem. But simply using task identity will be a problem, since there is no way to initialize the identity of the new task except random initialization. We try to solve this problem via a meta-encoder that learns the task representation from meta-information which is available on the new task as well.
Specifically, the meta-encoder is used to encode the task-specific information and generate the representation of the task as part of parameters. Through this way, different tasks will have different representations, thus different initial models. Also, since the presentation of the task is available, the model can leverage the relationship between different tasks.
To apply this idea in our problem, we propose two meta-encoder to encode two different kinds of task-specific information. One is to use the neighbor encoder to encode the start entity and the end entity, and then use the difference between the embedding of the start entity and the end entity as the task representation. But this take-specific information is not robust when the number of neighbors is small. Thus we propose another way for the case which encodes the path from the start entity to the end entity.
On two constructed few-shot multi-hop reasoning datasets, we show that the augmented meta-encoder yields much better initial point and outperforms several few-shot learning baselines.

The main contributions of this work include:

    $\bullet$ We introduce few-shot learning on the task of multi-hop reasoning over knowledge graph, and present two constructed datasets for this task.

    $\bullet$ We propose to use meta-encoder to encode task-specific information so as to generate better task-dependent model for the new task.
    
    $\bullet$ We apply neighbor encoder and path encoder to leverage the task-specific information in multi-hop reasoning task, and experiments verify the effectiveness of the augmented meta-encoder.

\section{Related Work}
\paragraph{Reasoning over Knowledge Graphs}
Knowledge graph reasoning aims to infer the existence of a query relation between two entities. There are two general approaches for knowledge graph reasoning. The embedding based approaches \cite{imt_nickel,imt_BordesUGWY13,imt_YangYHGD14a,imt_TrouillonWRGB16,kb_representaion_wu} learn the representations of the relations and entities in the KG with some heuristic self-supervised loss functions, while path search based approaches \cite{randomWalk_Lao,imt_NeelakantanRM15,DeepPath,MINERVA,variationalKGR,kgr_reward_reshaping,kgr_MCTS} solve this problem through multi-hop reasoning, i.e., finding the reasoning path between two entities. In spite of the superior performance of embedding-based methods, they can not capture the complex reasoning patterns in the KG and are lack of explainability. 

Due to its explainability, multi-hop reasoning has been investigated a lot in recent years. The Path-Ranking Algorithm (PRA) \cite{randomWalk_Lao} is a primal approach that learns random walkers to leverage the complex path features. \cite{GardnerTKM13,GardnerTKM14} improves upon PRA by computing feature similarity in the vector space. Recursive random walk integrates the background KG and text \cite{WangC15}. There are also other methods using convolutional neural network \cite{FB15K-237} and recurrent neural networks \cite{imt_NeelakantanRM15}. More recently, \cite{DeepPath} first applies reinforcement learning for learning relational paths. \cite{MINERVA} proposes a more practical setting of predicting end entity given the query relation and the start entity. \cite{kgr_reward_reshaping} reshapes the rewards using pre-trained embedding model. \cite{kgr_MCTS} uses Monte Carlo Tree Search to overcome the problem of sparse reward.

\paragraph{Meta-learning}
Meta-learning aims to achieve fast adaption on new tasks through meta-training on a set of tasks with abundant training examples. It has been widely applied in few-shot learning settings where limited samples are available \cite{MAML_translation,MAML_SQG}. 
One important category of meta-learning approaches is initialization based methods, which aims to find a good initial model that can fast adapt to new tasks with limited samples \cite{maml,reptile}. However, they only learn a single initial model and do not leverage the relationship between tasks. \cite{MateLatentEnbeddingOpt} proposes to learn a data-dependent latent generative representation of the model parameters and conduct gradient-based adaptation procedure in this latent space.
Another related work is Relation Network \cite{relationNet}, which consists of am embedding module to encode samples and a relation module to capture the relation between samples.

\section{Background}
In this section, we 
 will first introduce the multi-hop reasoning task. Then we will extend it to the meta-learning setting and introduce the popular framework (MAML) for few-shot learning.
\subsection{Multi-hop Reasoning Problem}
In this problem, there is a background graph $G$, and a set of query relations $R$. Each query relation has its own training and testing triple ($e_s,  r, e_t$), where $e_s$, and $e_t$ are the start entity and end entity in the KB, while $r$ is the query relation.  Given the start entity $e_s$ and the query relation $r$, the task is to predict the end entity $e_t$, along with a support reasoning path from $e_s$ to $e_t$ in $G$. The length of the path is set to be fixed, and an additional \emph{STOP} edge is added for each entity to point at itself so that the model is able to stay in the end entity.

We give an example to better explain this task. Consider the relation of \emph{Nationality} with a training triple: \emph{(Obama, Nationality, American)}. Given the start entity and the query relation, \emph{(Obama, Nationality)}, the model is expected to find a path with a fixed length in $G$ from \emph{Obama} to \emph{American}. A general framework to solve this problem is to train an agent that predicts the next relation based on the current entity, the query relation, and the visited path at each step. In expectation, the agent should give the reasoning path \emph{(BornIn, CityIn, ProvinceIn)}, and predict the end entity as \emph{American}.

\subsection{Meta-learning for Multi-hop Reasoning}
For multi-hop reasoning problem, we define a task as the inference of a specific relation's end entity conditioned on the start entity. It is easy to see that each relation forms an individual task. In the meta-learning framework, the tasks are divided into three disjoint sets called meta-training, meta-dev, and meta-test set respectively.
The goal of meta-learning is to train an agent that can quickly adapt on the new tasks in meta-test set with limited data by leveraging prior learning experience.

Following standard meta-learning setting as in \cite{maml}, our setting consists of two phases, the \textbf{meta-training} and \textbf{meta-test} phase.
In the meta-training phase, the agent learns on a set of meta-training tasks $\mathcal{T}=\{\mathcal{T}_1, \mathcal{T}_2,\cdots, \mathcal{T}_N\}$, where each task $\mathcal{T}_i$ has its own training and validation set denoted as \{$D_i^{train}$, $D_i^{valid}$\}. By learning on the meta-training tasks $\mathcal{T}$, the agent is expected to gain some knowledge about the reasoning process, which can help learn faster on new tasks.

In the meta-test phase, the trained agent will be evaluated on a set of new tasks in the meta-dev/meta-test task set $\mathcal{T}^\prime=\{\mathcal{T}_1^\prime, \mathcal{T}_2^\prime,\cdots, \mathcal{T}_{N^\prime}^\prime\}$. Each task $\mathcal{T}_i^\prime$ has its own training and testing set denoted as \{$D_i^{\prime train}$, $D_i^{\prime test}$\}, where $D_i^{\prime train}$ only has limited training samples. The agent will be fine-tuned on each task $\mathcal{T}_i^\prime$ using $D_i^{\prime train}$ for fixed gradient steps, and be evaluated after each gradient step. The macro-average on all tasks in $\mathcal{T}^\prime$ is reported as its performance of meta-learning. Note that the number of fine-tuning steps should be chosen according to the model's performance on meta-dev tasks, and use the fixed chosen steps on meta-test tasks directly, since there are only limited samples in the new task, which are not sufficient for choosing a feasible fine-tuning step.

\subsection{MAML Framework}
Let $f$ denotes the reasoning model in our setting that maps the observation to the action, i.e., next relation to be taken. The objective of MAML \cite{maml} is to find a good model initialization $f_\theta$ which can quickly adapt to the new tasks after a few adaptions. We will first introduce the objective function of MAML, and then illustrate how to optimize it in the following part. 

Let $\theta$ denote the parameter of the current model, and $\theta^\prime$ denote the updated parameter using samples from task $\mathcal{T}_i$. For example, suppose we use one gradient update on $\mathcal{T}_i$, then we have:
$$
\theta_{i}^{\prime}=\theta-\alpha \nabla_{\theta} \mathcal{L}_{\mathcal{T}_{i}}\left(f_{\theta}\right).
$$
The meta-objective is to optimize the performance of $f_{\theta_i^\prime}$ across tasks sampled from $p(\mathcal{T})$. More formal definition is as follows:
\begin{equation*}
\min _{\theta} \sum_{\mathcal{T}_{i} \sim p(\mathcal{T})} \mathcal{L}_{\mathcal{T}_{i}}\left(f_{\theta-\alpha \nabla_{\theta} \mathcal{L}_{\mathcal{T}_{i}}\left(f_{\theta}\right)}\right)
\end{equation*}

To optimize this problem, we sample a batch of tasks $
\mathcal{T}_i\sim p(\mathcal{T})$. For each task $\mathcal{T}_i$, two subsets ($D_i$ and $D_i^\prime$) of training examples will be sampled independently. $D_i$ is used to compute the updated parameters $\theta^\prime$. Then $\theta$ is optimized to minimize the objective function using $D_i^\prime$. Formally, we have
$$
\theta_{i}^{\prime}=\theta-\alpha \nabla_{\theta} \mathcal{L}_{\mathcal{T}_{i}}^{D_i}\left(f_{\theta}\right).
$$
$$
\theta = \theta - \beta \nabla_\theta \sum_{\mathcal{T}_i \sim p(\mathcal{T})}\mathcal{L}_{\mathcal{T}_i}^{D_i^\prime}(f_{\theta_i^\prime})
$$
The above optimization requires the computation of second-order gradient, which is computationally expensive. In practice, people usually use first-order update rule instead, which has similar performance but needs much less computation \cite{maml,reptile}:
$$
\theta = \theta - \beta \nabla_{\theta^\prime} \sum_{\mathcal{T}_i \sim p(\mathcal{T})}\mathcal{L}_{\mathcal{T}_i}^{D^\prime}(f_{\theta^\prime})
$$

\section{Meta-Learning of Deep Reasoners}

\subsection{MAML with Task-specific Initialization}
MAML learns a single initial model that does not depend on any task-specific information. It works by adapting the initial model through gradient update on the target task. In other words, the initial model learns some common knowledge shared by the tasks, so that it can adapt to new tasks quickly. However, MAML is not able to capture the relationship between different tasks because it is lack of task-specific information. One easy way to inject task information is to use task identity, such as the embedding of query relation in our KB reasoning problem. But this solution could incur two problems. First, the model will learn some knowledge that only applies to a specific task, which is hard to transfer when adapting to new tasks. Second, when there comes a new task, we can not easily initialize the task identity, e.g. the embedding of a new query. Therefore, we propose to use a meta-encoder to encode the task-specific information, which can not only enable the model to learn the relationship between different tasks but also allows the model adapt in the new task faster since the model can leverage the task-specific information of the new tasks as well.

Let $x$ and $\hat{x}$ denote the input data and task-specific information respectively. $g$ is the meta-encoder that encodes $\hat{x}$, and $f$ is the model which takes both $x$ and $g(\hat{x})$ as inputs to predict the outputs, i.e., $f(x, g(\hat{x}))$ is used for prediction.
Note that we hope $g(\hat{x})$ can encode the information about the whole target task instead of just $x$ itself so that $g(\hat{x})$ can also
benefit other instances $x^\prime$ within the same task $\mathcal{T}_i$, i.e., $f(x^\prime, g(\hat{x}))$ should perform well for any $x^\prime \in \mathcal{T}_i$. This is because the task-specific information may not be available for the testing sample. For example, the end entity we use as the task-specific information is not available in new testing samples.
To achieve this goal, we apply meta-gradient methods which is similar to MAML. Given a task $\mathcal{T}_i$, we will sample two subsets of instances $D_i$ and $D_i^\prime$. The updated parameter is computed using $D_i$:
$$
\theta_{i}^{\prime}=\theta-\alpha \nabla_{\theta} \mathcal{L}_{\mathcal{T}_{i}}\left(f_{\theta}(D_i, g_\theta(\hat{D}_i))\right).
$$
Then meta-gradient is computed using $\hat{D}_i$ and $D_i^\prime$, where $\hat{D}_i$ is used for initialization.
$$
\theta = \theta - \beta \nabla_\theta \sum_{\mathcal{T}_i \sim p(\mathcal{T})}\mathcal{L}_{\mathcal{T}_i}(f_{\theta_i^\prime}(D_i^\prime, g_{\theta_i^\prime}(\hat{D}_i)))
$$
The first order update rule can be written as:
$$
\theta = \theta - \beta \nabla_{\theta_i^\prime} \sum_{\mathcal{T}_i \sim p(\mathcal{T})}\mathcal{L}_{\mathcal{T}_i}(f_{\theta_i^\prime}(D_i^\prime, g_{\theta_i^\prime}(\hat{D}_i)))
$$

The details are shown in Algorithm \ref{algo_1}. At first, a batch of tasks will be sampled. For each task $\mathcal{T}_i$, we sample two subsets of instances $(D_i, D_i^\prime)$, and compute the meta information $\hat{D}_i$ based on $D_i$, which is the neighbor of start and end entity or the reasoning path between them for the multi-hop reasoning problem. In the following procedure, the updated parameters $\theta_i^\prime$ will be computed for each task (line 7-9). In meta-update step (line $11$), we update $\theta$ to minimize the loss of $\theta_i$ using new instances $D_i^\prime$ and the task representation $\hat{D}_i$.

For testing on a new task $\mathcal{T}^\prime_i$, we obtain the task representation $g(\hat{x})$ based on the few-shot samples $x\in D^{\prime train}_i$. Then we fine-tune $f$ and $g$ using the data $D^{\prime train}_i$. 
The model makes prediction on testing samples $x\prime \in D_i^{\prime test}$ using $f(x\prime, g(\hat{x}))$.

\begin{figure*}
    \centering
    \includegraphics[width=\linewidth]{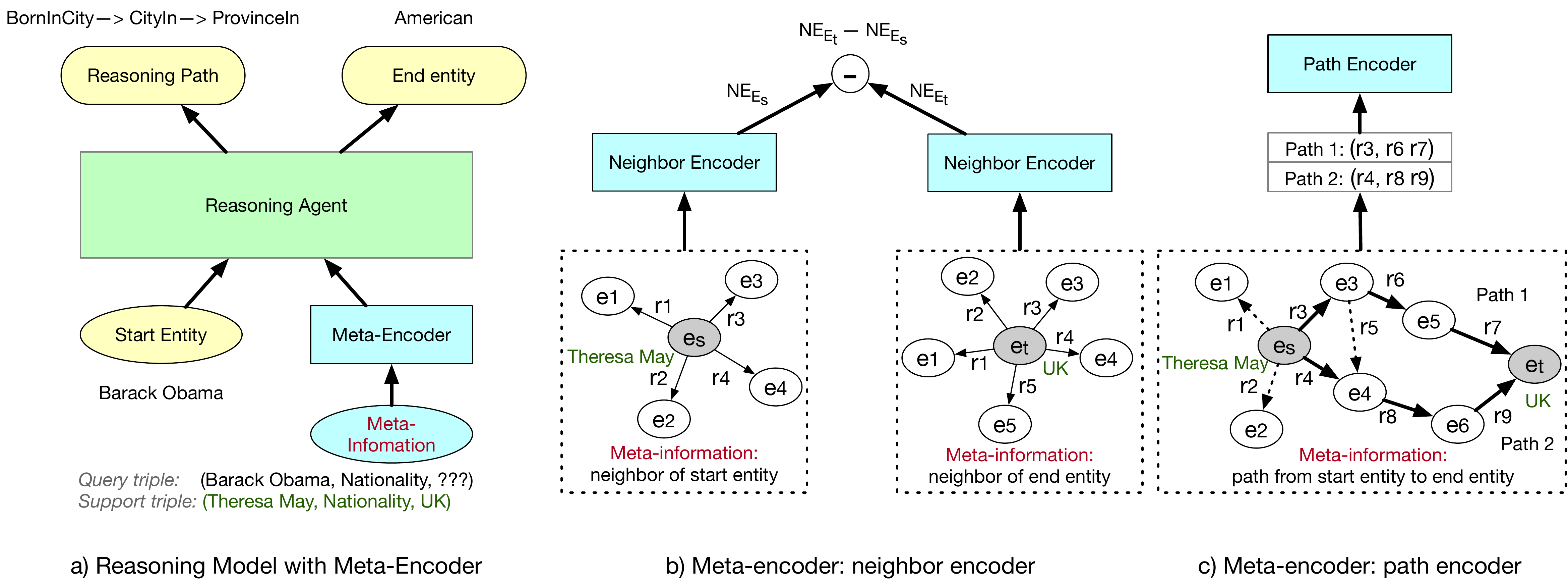}
    \caption{The model we use for meta-reasoning over knowledge graph. a) is the general framework of the model. b) and c) are our neighbor encoder and path encoder respectively.}
    \label{fig:model}
\end{figure*}

\begin{algorithm}[!t]
\small
\caption{MAML with Meta-Encoder}\label{algo_1}
\begin{algorithmic}[1]
    \REQUIRE ~~\\
    $p(\mathcal{T})$: the distribution of tasks\\
    $\alpha,\beta$: learning rates for adaptation and meta-update\\
    $k$: the number of adaptations\\
    $f, g$: the reasoning model and meta-encoder
    \STATE Randomly initialize $\theta$
    \FOR{step = 0 : M-1}
    \FOR{batch of tasks $\mathcal{T}_i\sim p(\mathcal{T})$}
    \STATE Sample task instances ($D_i, D_i^\prime$) from $\mathcal{T}_i$
    \STATE Compute task specific information $\hat{D}_i$
    \STATE Set $\theta_i^\prime = \theta$
    \FOR{i = 0 : k}
    \STATE $\theta_i^\prime \xleftarrow{} \theta_i^\prime-\alpha \nabla_{\theta_i^\prime} \mathcal{L}_{\mathcal{T}_{i}}(f_{\theta_i^\prime}(D_i, g_{\theta_i^\prime}(\hat{D}_i)))$
    \ENDFOR
    \ENDFOR
    \STATE $\theta \xleftarrow{} \theta - \beta \nabla_{\theta} \sum_{\mathcal{T}_i \sim p(\mathcal{T})}\mathcal{L}_{\mathcal{T}_i}(f_{\theta_i^\prime}(D_i^\prime, g_{\theta_i^\prime}(\hat{D}_i)))$
    \ENDFOR
\end{algorithmic}
\end{algorithm}

\subsection{Model}
The general framework of our model is shown in Figure \ref{fig:model}. The original reasoning agent takes start entity and query relation as inputs, and output the reasoning path and end entity. But this agent will not work well under meta-learning setting, where the embedding of the new query relation is hard to be initialized. Our method replaces the query relation with a meta-encoder that encodes some meta information about the task, which is available on a new task. In the following parts, we will introduce more about the reasoning agent and meta-encoder. 
\subsubsection{Reasoning Agent}
We use the policy proposed in \cite{MINERVA}, which is called MINERVA. They formulated this problem as a reinforcement learning problem. The state is defined as the combination of the query, the answer, and the current location (an entity in KB). But the answer is not observed, so the observation only includes the query and the current location. The actions are defined as the outgoing edges of the current location. The reward is $+1$ is reaching the answer, otherwise, it is $0$.

The policy uses LSTM to encode the history information, i.e. the visited path.
$$
\mathbf{h}_{\mathbf{t}}=\mathrm{LSTM}\left(\mathbf{h}_{\mathbf{t}-\mathbf{1}},\left[\mathbf{a}_{\mathbf{t}-\mathbf{1}} ; \mathbf{o}_{\mathbf{t}}\right]\right)
$$
where $\mathbf{h}_{\mathbf{t-1}}$ is previous hidden state, $\mathbf{a}_{\mathbf{t}-\mathbf{1}}$ is the embedding for the chosen relation at time $t-1$, and $\mathbf{o}_{\mathbf{t}}$ is the embedding of the current entity. The hidden state of the LSTM, $\mathbf{h}_{\mathbf{t}}$ is then concatenated with the embedding of the current entity $\mathbf{o_t}$ and the query relation $\mathbf{r_q}$. The action distribution $\mathbf{d_t}$ is computed by applying softmax on the matching score between the action embedding and the projection of the concatenated embedding, i.e.,
$$
\mathbf{d}_{\mathbf{t}}=\operatorname{softmax}\left(\mathbf{A}_{\mathbf{t}}\left(\mathbf{W}_{2} \operatorname{ReLU}\left(\mathbf{W}_{\mathbf{1}}\left[\mathbf{h}_{\mathbf{t}} ; \mathbf{o}_{\mathbf{t}} ; \mathbf{r}_{\mathbf{q}}\right]\right)\right)\right).
$$
The model structure is the same as proposed in \cite{MINERVA}, which uses two linear layers ($\mathbf{W_1}$ and $\mathbf{W_2}$) to encode the observation. Next action is sampled from the action distribution $\mathbf{d_t}$.

\subsubsection{Meta-encoder}
We can regard the embedding of query relation used in the above MINERVA model as the task identity. But when there comes a new task, there is no good way to find an initial embedding for the new query relation that fits into the reasoning model well. Therefore, we need another meta-encoder that leverage some meta-information about the new task and generate the embedding of query relation, based on which the model will be able to make reasonable outputs. Here we introduce two task-specific encoders to achieve this, neighbor encoder and path encoder.

\paragraph{Neighbor Encoder}
Given an instance, i.e., a triple $(e_s, r, e_t)$, we use the difference between the embedding of start entity $e_s$ and end entity $e_t$ as an representation of the query relation $r$ \shortcite{imt_BordesUGWY13}. To better represent the entity, we borrow the idea of neighbor encoder from \cite{one-shot-KBR}. Let $\mathcal{N}_e$ denotes the neighbor of entity $e$. For each relation-entity pair $(r_i, e_i)\in \mathcal{N}_e$, We compute the feature representation $C_{r_i, e_i}$ as
$$
C_{r_{i}, e_{i}}=W_{c}\left(v_{r_{i}} \oplus v_{e_{i}}\right)+b_{c},
$$
where $v_{r_i}$ and $v_{e_i}$ are the embedding for $r_i$ and $e_i$ respectively, $\oplus$ denotes concatenation, and $W_c$ and $b_{c}$ are parameters of a linear layer. Then the neighbor embedding of the given entity $e$ is computed as the average of the feature representations of all neighbors, i.e.,
$$
\text{NE}_{e} = \sigma (\frac{1}{\left|\mathcal{N}_{e}\right|} \sum_{\left(r_{i}, e_{i}\right) \in \mathcal{N}_{e}} C_{r_{i}, e_{i}} ),
$$
where $\sigma = \tanh$ is the activation function. Then the representation of the query relation is defined as the difference between the neighbor embedding of $e_s$ and $e_t$ like TransE \cite{imt_BordesUGWY13}:
$$
R_r = \text{NE}_{e_t} - \text{NE}_{e_s}.
$$

\paragraph{Path Encoder}
The neighbor encoder needs to encode the neighbor as the representation for the start and end entity, and it will not work well when the number of neighbors is small. Thus we propose another encoder for this case called path encoder.
Path encoder takes into consideration of the successful path in the graph, i.e., the reasoning path from start entity to end entity for a given query relation. Since not all the paths from start entity to end entity are meaningful, this path encoder is noisier than the neighbor encoder.

Let $\mathcal{P}_e$ denotes all the paths from start entity $e_s$ to end entity $e_t$. For any path $p_i \in \mathcal{P}_e$, we have $p_i= \{r_i^1, \cdots, r_i^n\}$, where $r_i^j$ is the selected relation at step $j$ in path $p_i$, and $n$ is the max length of reasoning path.
We use LSTM \cite{LSTM} to encode each path:
$$
\mathbf{h_t} = \text{LSTM}(\mathbf{h_{t-1}}, \mathbf{r_i^j}),
$$
where $\mathbf{h_t}$ is the hidden state of the LSTM at step $t$, and $\mathbf{r_i^j}$ is the embedding for relation $r_i^j$. The last hidden state $\mathbf{h_n}$ is used as the embedding $C_{p_i}$ for path $p_i$, i.e., $C_{p_i}=\mathbf{h_n}$. The final path embedding $\text{PE}_e$ for the given triple $(e_s, e, e_t)$ is average embedding of all the paths, i.e.,
$$
\text{PE}_e = \frac{1}{\left|\mathcal{P}_{e}\right|} \sum_{p_i \in \mathcal{P}_e} C_{p_i}.
$$

\begin{table}[!t]
    \centering
    \resizebox{0.48\textwidth}{!}{
    \begin{tabular}{c c c c c c c c c}\toprule
    \multirow{2}{*}{Dataset} & \multirow{2}{*}{\# Entities} & \multirow{2}{*}{\# Relations} & \multirow{2}{*}{\# Triples} & \multirow{2}{*}{\# Tasks} & \multicolumn{2}{c}{\# Degree} \\\cline{6-7}
    & & & & & average & median\\\midrule
    FB15K-237  & 14505 & 237 & 239266 & 237 & 20.00 & 14\\
    NELL   & 68272 & 358 & 181109 & 67 & 3.99 & 1 \\\bottomrule
    \end{tabular}
    }
    \caption{Statistics of the datasets. \# Entities, \# Relations, \# Triples, \# Tasks denotes the number of entities, relations, triples, tasks in the corresponding dataset respectively. In the column of \# Degree, average and median denote the average and median outgoing degree of each entity respectively.}
    \label{dataset_stat}
\end{table}

\begin{table*}[!t]
\small
\centering
\begin{tabular}{l| l | c c c c |c c c c}
\toprule
    \multirow{2}{*}{Setting} & \multirow{2}{*}{Method} & \multicolumn{4}{c|}{FB15K-237} & \multicolumn{4}{c}{NELL}\\
    & &Hits@1 & Hits@3 & Hits@10 & MRR & Hits@1 & Hits@3 & Hits@10 & MRR\\
    \midrule
    Full Data & MINERVA & .124 & .146 & .187 & .142 & .137 & .176 & .202 & .163 \\\midrule
    \multirow{8}{*}{Best}&&\multicolumn{8}{c}{\it Baselines}\\
    &Random & .017 & .028 & .043 & .027 & .047 & .100 & .165 & .086\\
    &Transfer & .010 & .012 & .054 & .019 & .041 & .070 & .128 & .066 \\
    &MAML & .021 & .041 & .052 & .035 & .067 & .086 & .139 & .087 \\
    &MAML-Mask & .009 & .023 & .045 & .019 & .032 & .054 & .080 & .058 \\\cline{2-10}
    &&\multicolumn{8}{c}{\it Ours}\\
    &Neighbor & \bf .065 & \bf .073 & \bf .128& \bf .080 & .045 & .066 & .106 & .064 \\
    &Path & .041 & .067 & .101 & .060 & \bf .108 & \bf .141 & \bf .200 & \bf .137\\\midrule
    \multirow{8}{*}{Initial}&&\multicolumn{8}{c}{\it Baselines}\\
    &Random & .000 & .000 & .005 & .002 & .021 & .074 & .105 & .056\\
    &Transfer  & .000 & .005 & .023 & .006 & .037 & .055 & .077 & .051 \\
    &MAML  & .005 & .005 & .023 & .010 & .017 & .031 & .054 & .032 \\
    &MAML-Mask  & .000 & .014 & .045 & .012 & .021 & .050 & .081 &  .043\\\cline{2-10}
    &&\multicolumn{8}{c}{\it Ours}\\
    &Neighbor  & \bf .043 & \bf .054 & \bf .092 & \bf .056 & .026 & .047 & .091 & .045\\
    &Path & .000 & .005 & .058 & .012 & \bf .082 & \bf .109 & \bf .164 & \bf .104 \\
\bottomrule
\end{tabular}
\caption{The results on $5$-shot experiments. We also report the performance of MINERVA on these tasks using full data for better comparison. Full Data denotes using MINERVA algorithm on these tasks with full training data. Best denotes the best performance for each method after fine-tuning, and Initial denotes the performance of method at the initial point. We report the average performance on the meta-test tasks. Best result for each evaluation matrix is marked in bold.}
\vspace{-0.2in}
\label{best_5_shot}
\end{table*}

\begin{figure*}[!t]
\centering
\begin{minipage}{0.49\linewidth}\centering
\includegraphics[width=7cm]{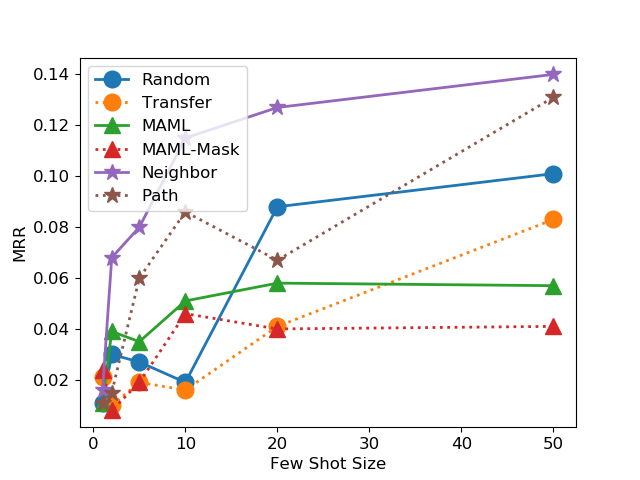}\\
(a) FB15K-237
\end{minipage}
\begin{minipage}{0.49\linewidth}\centering
\includegraphics[width=7cm]{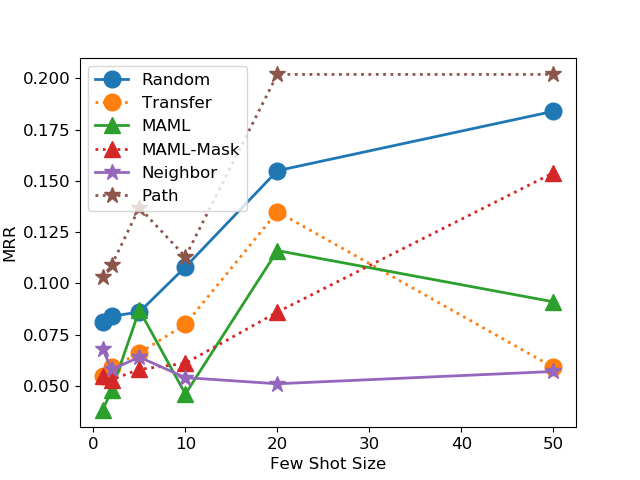}\\
(b) NELL
\end{minipage}
\caption{The change of the performance with the size of few-shot samples for each method. Here we choose the size to be $1, 2, 5, 10, 20, 50$. MRR of each model after fine-tuning is reported.}
\vspace{-0.2in}
\label{shot_size}
\end{figure*}

\section{Experiments}
To verify the effectiveness of the proposed methods, we compare it with several baselines on two knowledge completion datasets, FB15K-237 \cite{FB15K-237}, and NELL \cite{NELLDataset}. In the following part, we will introduce how we construct the meta-learning setting for knowledge graph reasoning and the baselines we use, then we will show the main results and other analytic experiments.

\subsection{Datasets and Settings}

We construct the meta-learning setting from two well-known knowledge completion datasets: FB15K-237 \cite{FB15K-237} and NELL \cite{NELLDataset}. FB15K-237 is created from original FB15K by removing various sources of test leakage. Every relation in the training set of FB15K-237 is regarded as an individual task. For the NELL dataset, we use the modified version from \cite{one-shot-KBR}, which chooses relations with more than $50$ triples, and less than $500$ triples as one-shot tasks. Here we used those selected tasks as meta-learning tasks. The statistics of the two datasets are shown in Table \ref{dataset_stat}.

Let $D_{\textrm{train}}$, $D_{\textrm{dev}}$, and $D_{\textrm{test}}$ denotes the training data, validation data and test data in original dataset such as FB15K-237. We choose some tasks with positive transfer (task that has better performance when training together with other tasks than training solely) as meta-dev and meta-test tasks. More specifically, we choose task with at least $0.04$ and $0.02$ positive transfer on FB15K-237 and NELL dataset respectively, from which we only keep tasks with more than $20$ samples in the dev set. Note that $0.04$ and $0.02$ are carefully chosen threshold so that we can get enough tasks with reasonable positive transfer. Through this way, we get $5/10$ and $4/7$ relations for meta-dev/meta-test on FB15K-237 and NELL respectively, and other relations left are used for meta-training. We denote the partitioned relation set as ${R_{\textrm{meta-train}}}/{R_{\textrm{meta-dev}}}/{R_{\textrm{meta-test}}}$, and each relation has its own training/test data.

\subsection{Baselines and Hyper-parameters}
We compare our methods with the following baselines. \textbf{Random} method trains a separate model for each task from random initialization. \textbf{Transfer} method will learn an initial model by using samples from $D_{\textrm{train}}^{\textrm{meta-train}}$. \textbf{MAML} uses the training framework of MAML to learn an initial point, and the task identity (the query relation) is given. \textbf{MAML-Mask} uses the same training framework as MAML, the difference is that we mask the task identity by setting the query relation for all tasks to be $0$. \textbf{Neighbor} and \textbf{Path} method means we use the neighbor encoder and path encoder to encode the task-specific information respectively.

We tuned the hyper-parameters for all the baselines and our methods, and they are set as follows. For Transfer, the batch size in the pre-training phase is set to be $128$. For MAML, MAML-Mask, Neighbor, and Path, the batch size is set to be $5$. For Path, $1$ adaption step is applied to compute the updated parameters, and $\alpha=0.01, \beta=0.001$. For Neighbor, MAML, and MAML-Mask, $5$ and $1$ adaption steps are applied on FB15K-237 and NELL respectively, and $\alpha=0.01$ when the number of adaption step $k=1$, $\alpha=0.001$ when $k=5$, and $\beta = 0.001$. Other parameters are set as default as in \cite{MINERVA}. 

\subsection{Results}
We conduct our experiments under $5$-shot learning setting, i.e., there are $5$ training samples for each task in ${R_{\textrm{meta-dev}}}$ and ${R_{\textrm{meta-test}}}$.
We use the mean reciprocal rank (MRR) and Hits@K to evaluate each model. For each method, we will first fine-tune and test the initial model on meta-dev tasks, through which we choose the number of fine-tune steps and fix it on meta-test tasks. For example, if a model has the best performance after $5$ fine-tune steps on meta-dev tasks, then the model will be tested after $5$ fine-tune steps on meta-test tasks. We report the best performance on meta-test tasks for each method in Table \ref{best_5_shot} as Best group. We also list the results using full data for better comparison. From the results, we can see that neighbor encoder and path encoder achieves the best performance on FB15K-237 and NELL dataset respectively. It is reasonable that neighbor encoder does not perform well on NELL dataset since the median outgoing degree on this dataset is only $1$. We also note that path encoder outperforms other baselines on FB15K-237, which verify the consistent effectiveness of the task-specific encoder. While other baselines do not show much difference as the simple Random baseline, sometimes they even underperform Random baseline.

In order to show that our model can have better initial point than others, we report the performance of the initial point without any training in Table \ref{best_5_shot} as Initial group. We notice that the baselines have very poor initial performances on FB15K-237, which is reasonable since the model has never seen the new relation.  From the results, we can see that the neighbor encoder and path encoder achieves much better initial point than other baselines in FB15K-237 and NELL respectively. The path encoder has a fair performance which is similar to the best of the baselines MAML-Mask, we think the reason that path encoder does not perform very well is the path encoder is noisier than neighbor encoder as we mentioned before.

\subsection{Few-shot Size}
To investigate the impact of the few-shot size on the performance of the model, we evaluate the model using various few-shot size: $1, 2, 5, 10, 20, 50$. The results are shown in Figure \ref{shot_size}. From the results, we can see that for MAML and MAML-Mask, their performances remain nearly the same after the size reaches $10$ on FB15K-237 dataset. The performance of MAML is not stable on NELL dataset, while MAML-Mask keeps increasing. Both methods underperform the Random baseline when the size increases. For Transfer method, its performance increases with the few shot size on FB15K-237, but there is a huge drop on NELL when the size is $50$, which indicates it is not stable enough, and sensitive to the noise in the data. The neighbor encoder has the best performance on FB15K-237 dataset, but not well on NELL due to the small neighbor size. Path encoder seems to be less stable compared with neighbor encoder since there is performance drop once on both datasets, but it achieves the best performance on NELL and second-best performance when size is larger than $5$ except $20$.

\begin{table}[!t]
\small
\centering
\begin{tabular}{l | c c c c}
\toprule
      \multirow{2}{*}{Setting} & \multicolumn{4}{c}{FB15K-237} \\
    &Hits@1 & Hits@3 & Hits@10 & MRR \\
    \midrule
    Encoder-1-shot & .047 & .058 & .117 & .064 \\
    Encoder-50-shot & \bf .049 & \bf .070 & \bf .128 & \bf .069 \\
    No-encoder & .008 & .035 & .084 & .032 \\
\bottomrule
\end{tabular}
\caption{The comparison of performance for model with different initialization on FB15K-237 dataset. Encoder-1-shot and Encoder-50-shot denotes using neighbor encoder with $1$ and $50$ samples. No-encoder means using a random initialization. We report the average performance on meta-test tasks. Best result for each evaluation matrix is marked in bold.}
\label{table_ablation_study}
\end{table}

\subsection{Ablation Study}
To verify the effectiveness of the encoder, we compare the model using task-specific initialization with the model using random initialization at the initial point. We choose the neighbor encoder on FB15K-237 dataset to conduct the ablation study.
The comparison results are shown in Table \ref{table_ablation_study}. The three models in the table use the same reasoning model, the only difference is the task representation. Encoder-1-shot and Encoder-50-shot apply neighbor encoder to generate the task representation using $1$ and $50$ samples respectively, while No-encoder uses a randomly initialized representation. By comparing Encoder-1-shot with No-encoder, we can see that the model can achieve much better performance through the way of encoding task-related information, even using only one sample, which also indicates the generated task representations are meaningful. Also, better initialization can be achieved when using more samples, since the performance of Encoder-50-shot is better than that of Encoder-1-shot.


\section{Conclusion}
In this paper, we consider multi-hop reasoning over knowledge graphs under few-shot learning setting, where limited samples are available on new tasks. 
We improve upon MAML by using a meta-encoder to encode task-specific information. Through this way, our method can create a task-dependent initial model that better fits the target task.
Neighbor encoder and path encoder are proposed for our problem. Experiments on FB15K-237 and NELL under meta-learning setting show that our task-specific meta-encoder yields a better initial point and outperforms other baselines.




\bibliography{emnlp-ijcnlp-2019}
\bibliographystyle{acl_natbib}


\end{document}